%% file: main.tex
\documentclass[letterpaper]{article}
\usepackage{booktabs}
\usepackage{aaai2026}
\nocopyright
\usepackage{times}
\usepackage{helvet}
\usepackage{courier}
\usepackage{amsmath,amssymb}
\usepackage[hyphens]{url}
\usepackage{graphicx}
\usepackage{natbib}
\usepackage{caption}

\usepackage{placeins}
\title{Self-Specialized Teachers for Domain Post-Training}

\author{
Yifei Li$^{1,2,*}$, Rongman Xu$^{1,2,*}$, Lingling Zhang$^{1,2,\dagger}$,
Muye Huang$^{1,2}$, Zihan Ma$^{1,2}$, Jiashuai Liu$^{1,2}$,
Hang Yan$^{1,2}$, Heng Wang$^{1,2}$
}

\affiliations{
$^{1}$School of Computer Science and Technology, Xi'an Jiaotong University\\
$^{2}$MOE KLNN Lab, Xi'an Jiaotong University\\
$^{*}$Equal contribution. \quad $^{\dagger}$Corresponding author.\\
\texttt{yifeilee@stu.xjtu.edu.cn}
}

\begin{document}
\maketitle

\begin{abstract}
Target-only post-training can improve performance in a specialized domain while degrading behaviors that a general-purpose base model acquired before adaptation. We study this problem when target-domain data are available but a representative replay corpus is not. We propose self-specialized teacher distillation (SSTD), a two-stage procedure that first trains a copy of the base model into a domain teacher, then distills its token distribution to a student on prefixes sampled from the student itself. Teacher training combines standard target supervision with base-aware key-token weighting and distribution alignment to the frozen base model; on-policy distillation then places domain feedback on states the student can encounter at inference time. On financial numerical reasoning, medical question answering, and legal holding identification, SSTD retains much of the target improvement of direct fine-tuning while improving the mean score on the evaluated general suite by 4.8--5.0 points at the reported operating point. The pattern persists across Qwen3 sizes and on Gemma backbones. SSTD requires neither an external teacher nor general replay data.
\end{abstract}


\section{Introduction}

\begin{figure}[t]
\centering
\includegraphics[width=0.98\linewidth]{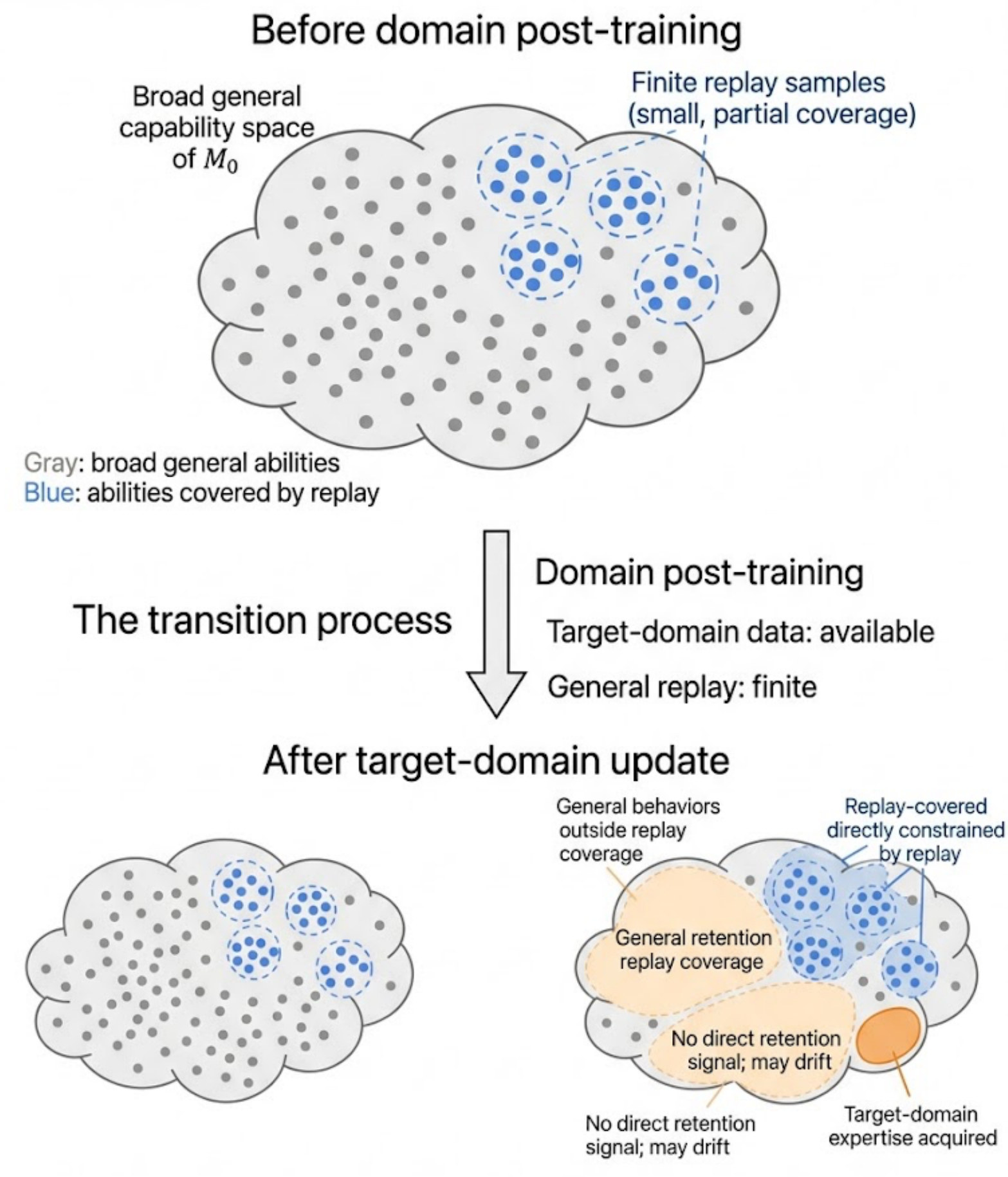}
\caption{A finite replay set directly constrains only a small, partial subset of a base model's general behavior. During target-domain post-training, general behaviors outside that coverage receive no direct retention signal and may drift, even as the model gains the desired domain expertise.}
\label{fig:intro}
\end{figure}
Scaling has turned pretrained language models into reusable general-purpose systems rather than starting points for a single downstream task. Large models transfer knowledge across distributions more effectively than models trained only on the downstream data~\citep{hernandez2021scaling,brown2020language}, and instruction and preference post-training expose broad behaviors such as reasoning and instruction following~\citep{wei2022flan,ouyang2022instructgpt}. A later domain adaptation inherits all of those behaviors, including many that its target benchmark does not name or measure. They remain useful when a financial assistant must follow an unfamiliar instruction, when a medical answer requires ordinary reading comprehension, or when a legal response depends on general reasoning before it reaches a domain-specific decision. As this asset grows, domain post-training cannot be judged only by its target score: it must add the desired domain behavior without needlessly changing useful behavior outside that domain. Prior studies document losses in general knowledge and reasoning after LLM fine-tuning~\citep{luo2023empirical,liu2024gci}. We study this problem as controlled domain adaptation, rather than as a sequence of tasks in the usual continual-learning protocol. The question is not whether a model can be made more specialized, but how narrowly an update can add specialization to a broadly capable model.

Our setting is deliberately restrictive but common in practice. A practitioner has target-domain data and a capable base model, but not the broad pretraining corpus that created the model. They may retain a finite replay set, yet such a set can directly constrain only the behaviors it happens to cover; it cannot stand in for an unknown and much wider distribution of prompts, reasoning patterns, and response styles. Figure~\ref{fig:intro} illustrates the resulting asymmetry. Target-domain data provide a direct signal for the desired expertise, whereas the much larger space of general behaviors receives no such signal outside the replay-covered regions. Target-only SFT can therefore acquire a domain skill while moving unrelated responses; replay and reference matching reduce that movement only where their examples or constraints apply~\citep{chaudhry2019agem,rebuffi2017icarl,li2018lwf}. Parameter penalties face a related limit: they restrict change without saying which changes carry the new domain knowledge. Existing continual-learning methods address retention across a sequence of earlier tasks, with task order and accumulated drift as central variables. We instead ask what a single focused update does to a general-purpose model when broad replay is unavailable. The problem is not to preserve every aspect of the base model unchanged. It is to identify the domain changes that matter, then make those changes without treating the rest of the model's behavior as expendable.

We start from a different view of the teacher. A useful teacher for post-training need not be the model with the highest standalone target-domain score. It should instead represent a domain increment that the student can absorb: one that corrects the base model's local gaps while remaining connected to its starting distribution. A target response usually mixes ordinary language, familiar response structure, and a smaller number of domain facts, calculations, or decisions. Uniform SFT treats all of these tokens alike, even though the base model may already handle much of the first two categories. This view motivates a \emph{self-specialized teacher}. We initialize the teacher from the same base model as the student, specialize it using the target data, and give extra weight to reference tokens that the frozen base model finds unlikely. These positions expose where target responses demand behavior the base model does not already supply, so the teacher spends more of its capacity on the student's actual domain gaps rather than simply refitting its existing behavior. The teacher must still remain teachable. A domain expert that assigns nearly all of its probability to tokens the student regards as impossible can have a good offline target score but yield a poor distillation signal. We therefore constrain the teacher distribution against the frozen base model while allowing the target-domain losses to move it when the data support a change. The constraint does not ask the teacher to copy the base model; it asks the teacher to express its new expertise as a correction that remains locally accessible to the student. The result is neither a generic external expert nor a teacher trained only to preserve the base. It is a constrained domain expert whose learned change has a path back to the student. This design follows the broader observation that a teacher--student gap can obstruct distillation~\citep{mirzadeh2020teacherassistant}, while using the shared initialization that makes self-distillation useful even between models of equal architecture~\citep{furlanello2018born}.

That domain increment must also reach the student at states it will encounter at inference time. Fixed gold responses and teacher trajectories condition on ideal prefixes; a student instead continues from its own earlier predictions, including locally plausible mistakes that alter every following state. A teacher answer on the gold path gives no direct guidance after such a deviation. We therefore let the student generate each trajectory and query the frozen self-specialized teacher at the resulting prefixes. The student chooses the states to train on, whereas the teacher supplies a full next-token distribution at each state. This on-policy stage transfers domain guidance where the student needs it, rather than only along an offline response path~\citep{agarwal2024onpolicy}. We call the resulting procedure \emph{self-specialized teacher distillation} (SSTD). Across financial numerical reasoning, medical question answering, and legal holding identification, SSTD retains most of the target gain of direct SFT while reducing the loss on held-out general evaluations. Ablations separate the contribution of the self-specialized teacher from that of student-generated prefixes; experiments across Qwen3 scales and Gemma models test whether the resulting target--retention pattern transfers beyond one model size or family. SSTD requires neither an external teacher nor a broad general replay corpus.

\section{Related Work}
\paragraph{Domain adaptation and retention.}
Continued pretraining and target supervised fine-tuning are standard routes for adapting a pretrained language model to a domain; work on BioBERT, SciBERT, Legal-BERT, and domain-adaptive pretraining reports gains from in-domain text~\citep{lee2020biobert,beltagy2019scibert,chalkidis2020legalbert,gururangan2020dont}. These objectives, however, optimize behavior on the target distribution and do not directly test what remains outside it. Retention methods in continual learning take several forms. Episodic-memory methods revisit stored examples from earlier data~\citep{lopezpaz2017gem,chaudhry2019agem,rebuffi2017icarl}; pseudo-replay methods generate earlier-task examples~\citep{sun2020lamol}; parameter-based methods restrict changes to parameters deemed important for prior behavior~\citep{kirkpatrick2017ewc,zenke2017si,aljundi2018mas}; and learning-without-forgetting matches predictions of a frozen old model while training on new-task data~\citep{li2018lwf}. Fine-tuned language models have also been studied in explicit continual-learning protocols~\citep{scialom2022continual}. We study a narrower setting: one focused domain update, no broad replay corpus, and retention measured on held-out general evaluations. The teacher-alignment term in our method is closest in spirit to reference-model regularization, but it constrains the specialized teacher before it supervises the student.

\paragraph{Knowledge distillation and teacher construction.}
Knowledge distillation predates current language models and has included logit matching, intermediate representation matching, and teacher--student chains~\citep{bucilua2006model,ba2014deep,hinton2015distilling,romero2015fitnets,mirzadeh2020teacherassistant}. For sequence generation, sequence-level distillation commonly uses teacher-produced sequences as fixed training targets~\citep{kim2016sequence}; recent LLM distillation has also used rationales or reverse-KL objectives~\citep{hsieh2023stepbystep,gu2024minillm}. Distillation need not require a smaller student: born-again networks and deep mutual learning show that peers or identically parameterized models can provide useful supervision~\citep{furlanello2018born,zhang2018mutual}. Our teacher likewise shares the student's initialization and architecture, but it has a different purpose. It is first specialized with target-domain data, then constrained to remain locally compatible with the frozen base distribution; the student learns from this constrained domain increment rather than from a generic stronger teacher.

\paragraph{Distillation at student-generated states.}
Fixed gold or teacher trajectories need not cover prefixes produced by an autoregressive student after an early error. Scheduled sampling, Professor Forcing, and sequence-level training each address exposure to generated states or sequence-level generation objectives~\citep{bengio2015scheduled,lamb2016professor,ranzato2016sequence}. DAgger established the broader imitation-learning principle of querying an expert on states induced by the learner~\citep{ross2011dagger}. More recently, MiniLLM and on-policy language-model distillation optimize teacher feedback on generated sequences, with the latter explicitly training on student trajectories~\citep{gu2024minillm,agarwal2024onpolicy}. SSTD adopts this student-state supervision, but differs in the teacher it queries: our teacher is a target-specialized copy of the base model, trained with base-aware key-token guidance and distribution alignment. The experiments therefore separate the value of on-policy prefixes from the value of this teacher construction.

\section{Problem Setting}
We consider a base model $M_0$, target-domain training data $D_T$, and a student $M_S$ initialized from $M_0$. The training process may inspect $D_T$ and samples from $M_S$, but it does not receive a representative replay set from the base model's original pretraining distribution. The desired update improves a target-domain measure $Q_T$ while keeping losses on a held-out general evaluation suite $Q_G$ small. This definition deliberately separates the task from continual-learning sequences: there is one focused update, one target distribution, and a question about collateral change in a broadly capable model.

A useful evaluation reports both axes. Target scores alone cannot show whether adaptation damaged unrelated behavior, while a retention score alone cannot show whether the method learned the target domain. We therefore compare each trained model with the untouched base model on FinQA execution accuracy, MedMCQA accuracy, and CaseHOLD accuracy, and report changes on a held-out general suite containing MMLU, HellaSwag, and ARC-C. Our primary retention measure is the arithmetic mean of those three general-benchmark scores; Appendix~\ref{app:implementation-details} reports the individual scores.

\section{Self-Specialized Teacher Distillation}
\begin{figure*}[t]
\centering
\includegraphics[width=0.98\textwidth]{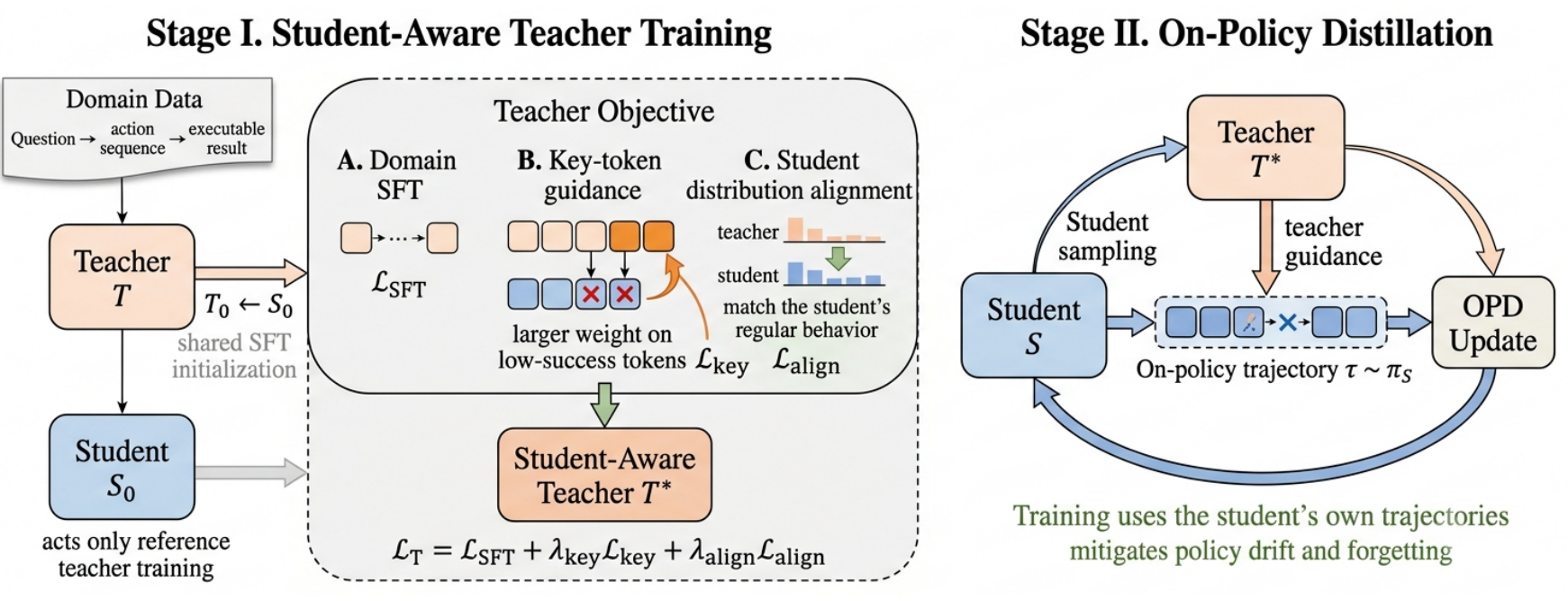}
\caption{Student-aware teacher training with on-policy distillation. In Stage~I, a copy of the base model specializes to the domain while remaining close to the frozen student distribution. In Stage~II, the student samples its own trajectories and receives dense token-level supervision from the resulting teacher.}
\label{fig:method}
\end{figure*}

An OPD teacher needs two properties that standard teacher selection treats separately: it must know the target domain, and its distributional change must remain accessible to the student it supervises. Figure~\ref{fig:method} separates these requirements into two stages. Let $S_0$ denote the frozen base model and let $D=\{(x,y)\}$ be target-domain examples, where $y=(y_1,\ldots,y_L)$ is the reference completion. A common choice is an independently trained, stronger teacher. That choice can improve the teacher's own likelihood or task score, yet it gives no guarantee that the teacher's preferred tokens lie near the student's current decision boundary. When the gap is large, distillation asks the student to imitate probability mass that it almost never assigns; the resulting supervision can be poorly conditioned even when the teacher response is correct.

We instead construct the expert from the student itself. We initialize $T$ from $S_0$ and train it only on $D$, while keeping the second copy frozen as a reference. This shared initialization gives $T$ a precise role: it represents an expert-domain increment over the behaviors already available to $S_0$, rather than an unrelated collection of capabilities. It also respects the setting in which the desired expert capability must arise through post-training of the model under study, without an external model or a broad replay corpus. The remaining issue is where that increment should be placed. Plain SFT would teach $T$ every target token equally, including many tokens that $S_0$ already handles; Stage~I therefore trains a teacher that concentrates on the student's local domain gaps while limiting departures that would make those corrections hard to transfer.

\subsection{Base-Aware Key-Token Guidance}
Ordinary supervised fine-tuning gives every reference token the same training role. That objective teaches $T$ to fit the domain on average, but it does not distinguish a token that $S_0$ already predicts well from one that exposes a local capability gap. Domain responses often contain both: familiar scaffolding, formatting, and common vocabulary coexist with a small set of domain facts, operations, or decisions that cause the base model to fail. Uniform training allocates teacher capacity according to token frequency, not according to which positions will later need correction. We measure the latter quantity with the frozen base model's token-level negative log-likelihood,
\begin{equation}
\ell_t^{\mathrm{base}}=-\log p_{S_0}(y_t\mid x,y_{<t}).
\label{eq:base-difficulty}
\end{equation}
We then convert the difficulty into a normalized weight,
\begin{equation}
w_t=(1-\alpha)+\alpha\frac{\ell_t^{\mathrm{base}}}
{\frac{1}{L}\sum_{j=1}^{L}\ell_j^{\mathrm{base}}},
\qquad \alpha\in[0,1].
\label{eq:key-weight}
\end{equation}
The score in Eq.~\ref{eq:base-difficulty} has a direct interpretation on a gold response: a high value means that the reference next token sits in a region where the frozen base has little confidence. We normalize within each response because absolute negative log-likelihood varies with response length, prompt format, and the overall difficulty of an example. The normalization keeps the mean weight equal to one, so varying $\alpha$ redistributes teacher capacity rather than changing the overall scale of the example loss. When $\alpha=0$, every position receives the same weight and the objective reduces to standard SFT. As $\alpha$ increases, tokens that the base assigns low probability receive more teacher-training signal. Since $S_0$ remains frozen, these weights are fixed throughout teacher training and do not introduce a moving target. The resulting key-token loss is
\begin{equation}
\mathcal{L}_{\mathrm{key}}=-\frac{1}{L}\sum_{t=1}^{L}w_t
\log p_T(y_t\mid x,y_{<t}).
\label{eq:key-loss}
\end{equation}
The reweighting does not discard the rest of a response: $w_t$ remains positive for every token, and the ordinary SFT term below still preserves full-sequence domain fitting. It changes the teacher's priority. $T$ no longer seeks only high average likelihood on $D$; it allocates extra modeling effort to the parts of target responses that reveal where $S_0$ falls short. This distinction matters in the next stage, where teacher probability mass becomes the student's update signal rather than an endpoint metric for the teacher alone.

\subsection{Student-Distribution Alignment}
Key-token reweighting deliberately pushes the teacher toward positions where the base is uncertain. That pressure alone can create a teacher that succeeds on the target data by making very sharp changes to its next-token distribution. Such a teacher may look stronger under an offline evaluation, but it can still be a poor tutor: if $T$ assigns high probability to tokens that $S_0$ regards as nearly impossible, the student receives a steep correction at each queried prefix and may fail to generalize that correction beyond the observed context. This is a teacher--student compatibility problem, not a lack-of-domain-knowledge problem.

We address it during teacher training rather than asking the student to resolve it later. We retain an ordinary domain-SFT term,
\begin{equation}
\mathcal{L}_{\mathrm{SFT}}=-\frac{1}{L}\sum_{t=1}^{L}
\log p_T(y_t\mid x,y_{<t}),
\label{eq:teacher-sft}
\end{equation}
and, on the same target-domain prefixes, constrain the teacher distribution against the frozen base distribution:
\begin{equation}
\mathcal{L}_{\mathrm{align}}=\frac{1}{L}\sum_{t=1}^{L}
\mathrm{KL}\!\left(p_T(\cdot\mid x,y_{<t})\,\middle\|\,
p_{S_0}(\cdot\mid x,y_{<t})\right).
\label{eq:teacher-align}
\end{equation}
The direction of the KL penalty matters. $\mathrm{KL}(p_T\|p_{S_0})$ charges the teacher when it puts substantial mass on a token that the base assigns little mass, which directly discourages unsupported teacher preferences. It does not require $T$ to reproduce every base-model probability; the SFT and key-token terms may still move the teacher toward target-domain behavior when the evidence warrants it. This term therefore has a narrow function. It does not supply domain knowledge or identify a student weakness. Instead, it keeps the learned expert increment close enough to the frozen student's regular behavior that the increment can later transfer through distillation.

\subsection{Student-Aware Teacher Objective}
We obtain the student-aware teacher by optimizing
\begin{equation}
\mathcal{L}_{T}=\mathcal{L}_{\mathrm{SFT}}
+\lambda_{\mathrm{key}}\mathcal{L}_{\mathrm{key}}
+\lambda_{\mathrm{align}}\mathcal{L}_{\mathrm{align}},
\label{eq:teacher-objective}
\end{equation}
where $\lambda_{\mathrm{key}}$ controls the contribution of base-aware key-token guidance and $\lambda_{\mathrm{align}}$ sets the penalty for departing from the frozen base distribution. These are the two loss coefficients shown in Figure~\ref{fig:method}. The reweighting parameter $\alpha$ remains internal to $w_t$ in Eq.~\ref{eq:key-weight}: it redistributes the token-level SFT signal but does not scale $\mathcal{L}_{\mathrm{key}}$. A small $\lambda_{\mathrm{key}}$ leaves the teacher close to an average domain model, whereas an excessively large value can over-focus teacher capacity on isolated hard tokens. Similarly, a small $\lambda_{\mathrm{align}}$ permits a sharper but less compatible expert, while a large value can suppress domain changes supported by the target data. We optimize only the parameters of $T$ in Eq.~\ref{eq:teacher-objective}; $S_0$ supplies a fixed diagnostic for both the difficulty weights and the compatibility term.

Each loss has a separate purpose in this construction. $\mathcal{L}_{\mathrm{SFT}}$ teaches the domain, $\mathcal{L}_{\mathrm{key}}$ selects where the teacher should exceed the student, and $\mathcal{L}_{\mathrm{align}}$ restricts how far that excess may move from the student's distribution. After optimization, we freeze $T^*$. It contains additional target-domain behavior, but its output distribution retains a connection to the model that will receive its logits in Stage~II.

\begin{table*}[t]
\centering
\normalsize
\renewcommand{\arraystretch}{1.16}
\setlength{\tabcolsep}{4.5pt}
\begin{tabular*}{\textwidth}{@{\extracolsep{\fill}}llcccccc}
\toprule
& & \multicolumn{2}{c}{Finance} & \multicolumn{2}{c}{Medical} & \multicolumn{2}{c}{Legal} \\
Model & Method & Target $\uparrow$ & General $\uparrow$
& Target $\uparrow$ & General $\uparrow$
& Target $\uparrow$ & General $\uparrow$ \\
\midrule
\raisebox{-25pt}[0pt][0pt]{Qwen3-1.7B}
& Base & 34.2 & 57.8 & 63.2 & 58.1 & 69.4 & 59.3 \\
& Naive SFT & \textbf{61.5} & 50.4 & \textbf{75.8} & 51.7 & \textbf{82.0} & 53.5 \\
& OPSD & 56.8 & 55.0 & 72.6 & 55.8 & 78.5 & 57.2 \\
& OPD & 55.5 & 53.0 & 71.8 & 54.3 & 77.6 & 55.6 \\
& \textbf{SSTD (ours)} & 59.5 & \textbf{55.2}
& 74.2 & \textbf{56.7}
& 80.1 & \textbf{58.4} \\
\bottomrule
\end{tabular*}
\caption{Domain adaptation and retention across domains. Target is FinQA
execution accuracy, MedMCQA accuracy, or CaseHOLD accuracy, respectively.
General is the mean of MMLU, HellaSwag, and ARC-C.}
\label{tab:main}
\end{table*}

\subsection{On-Policy Distillation}
Even a compatible teacher cannot remove the mismatch between gold response prefixes and the prefixes encountered at generation time. Offline SFT and offline distillation condition on $y_{<t}$, which belongs to the data distribution. An autoregressive student instead conditions on its own earlier choices. One locally plausible but incorrect token changes the subsequent context; after that point, training only on the gold path says little about the state at which the student must choose its next action. This discrepancy compounds over a long response, particularly when the target task requires multi-step reasoning or a structured output.

Stage~II places teacher supervision on the student's state distribution. Starting from a prompt $x$, the current student $S$ samples a trajectory $\tau=(\hat y_1,\ldots,\hat y_K)\sim\pi_S(\cdot\mid x)$. At each student-reached prefix $(x,\hat y_{<t})$, the frozen teacher $T^*$ conditions on exactly that prefix and supplies its full vocabulary distribution. The teacher does not choose the trajectory being optimized; the student does. This separates the two roles: student rollouts determine which states receive supervision, while the specialized teacher ranks possible next tokens at those states. We update the student with token-level distillation,
\begin{equation}
\begin{aligned}
\mathcal{L}_{\mathrm{OPD}}
&=\mathbb{E}_{\tau\sim\pi_S}\Biggl[\frac{1}{K}\sum_{t=1}^{K}\\[-2pt]
&\hspace{35pt}\operatorname{KL}\!\left(p_{T^*}(\cdot\mid x,\hat y_{<t})
\,\middle\|\,p_S(\cdot\mid x,\hat y_{<t})\right)\Biggr].
\end{aligned}
\label{eq:opd}
\end{equation}
Using full logits rather than a sampled teacher continuation retains information about alternatives that a single teacher sample would hide. For example, the distribution can give a high score to a correct operation while still assigning some mass to a syntactically valid fallback; a sampled target reduces that preference ordering to one token. The KL loss consequently provides dense feedback at every queried state and remains defined even when the student has already departed from the reference response. Because the states come from $S$ itself, the optimization targets prefixes that the student is likely to encounter at inference time. The final student starts from $S_0$ and updates under Eq.~\ref{eq:opd}, while $T^*$ stays fixed throughout Stage~II.

\section{Experiments}

\subsection{Experimental Setup}
\paragraph{Tasks and data.}
We study three independent domain-adaptation tasks: financial numerical
reasoning, medical question answering, and legal holding identification. We
use FinQA~\citep{chen2021finqa}, MedMCQA~\citep{pal2022medmcqa}, and
CaseHOLD~\citep{zheng2021casehold} with its LexGLUE split
\citep{chalkidis2022lexglue}, respectively. Every method trains only on the
official training split; validation data select hyperparameters and checkpoints,
and the held-out test split is evaluated once. The train/validation/test sizes
are 6,251/883/1,147 for FinQA, 182,822/4,183/6,150 for MedMCQA, and
45,000/3,900/3,900 for CaseHOLD. FinQA uses execution accuracy over the
generated reasoning program, whereas MedMCQA and CaseHOLD use multiple-choice
accuracy.

\paragraph{Models and training protocol.}
Qwen3-1.7B is the primary backbone in the multi-domain comparison. We then
repeat the Finance experiment with Qwen3-0.6B and Qwen3-4B, and with
Gemma-3-1B and Gemma-3-4B. Every run starts from its corresponding pretrained
checkpoint; no run continues from a checkpoint adapted to another domain. The
compared methods receive the same target-domain examples, maximum sequence
length, update budget, and decoding format.

\paragraph{Baselines and metrics.}
We compare direct supervised fine-tuning (Naive SFT), base-anchored
optimization (OPSD), offline distillation from an external teacher (OPD), and
SSTD. Target scores are compared within, rather than across, domains because
the target metrics differ. To measure retained general ability, we report the
arithmetic mean of MMLU~\citep{hendrycks2021mmlu},
HellaSwag~\citep{zellers2019hellaswag}, and ARC-C~\citep{clark2018arc}:
\[
\mathrm{General\ Avg.} =
\frac{s_{\mathrm{MMLU}} + s_{\mathrm{HellaSwag}} + s_{\mathrm{ARC-C}}}{3}.
\]
\paragraph{Implementation details.}
Appendix~\ref{app:implementation-details} provides the exact source checkpoints, teacher
initialization, optimizer schedule, sampling configuration, and the selected
values of $\alpha$, $\lambda_{\mathrm{key}}$, and $\lambda_{\mathrm{align}}$,
along with prompt templates, filtering rules, seed protocol, and compute accounting. It
also records the checkpoint-selection rule and per-seed results when repeated
runs are available.

\subsection{Domain Adaptation and General Retention}
Table~\ref{tab:main} compares domain adaptation and retained general
performance for Qwen3-1.7B. Naive SFT attains the highest target score in each
domain, but it also incurs the largest drop on the general suite. SSTD retains
most of the target gain: relative to Naive SFT, its target score is lower by
2.0 points on Finance, 1.6 points on Medical, and 1.9 points on Legal. In
return, it improves General Avg. by 4.8, 5.0, and 4.9 points, respectively.
For the reported domains, SSTD occupies a more favorable target--retention
trade-off than OPSD and OPD. The recovery on General Avg. is tightly grouped
(4.8--5.0 points) despite the three domains having different target tasks. This
result is consistent with the proposed training objective, although it does not
establish that every target domain will exhibit the same pattern.

The paired scores make the cost of that retention concrete. On Finance, SSTD
gives up 2.0 target points relative to Naive SFT (59.5 versus 61.5), while its
General Avg. is 4.8 points higher (55.2 versus 50.4). Medical and Legal show
the same ordering: the target gaps are 1.6 and 1.9 points, whereas the general
scores recover by 5.0 and 4.9 points. Thus, SSTD does not merely select a more
conservative checkpoint. It retains a substantial target improvement over the
base model while changing the target--retention balance relative to target-only
training.

\subsection{Teacher Objective Ablation}
Table~\ref{tab:teacher-ablation} isolates the three terms used to construct the
self-specialized teacher. Every row keeps the student rollout policy, student
objective, data budget, and update budget fixed; only the teacher objective
varies. Key-token guidance contributes the larger target
gain (+1.0 versus +0.6), whereas alignment gives the larger General Avg.
recovery (+1.1 versus +0.6). The split is informative: token-level guidance
helps the specialized task more, while alignment primarily limits the loss on
the general suite. Combining both terms yields the best result.

Neither term alone reproduces the full result. Key-token guidance raises the
target score to 59.0 but leaves General Avg. at 54.2, whereas distribution
alignment reaches 54.7 on General Avg. with a smaller target gain of 0.6. The
combined objective reaches 59.5 target and 55.2 general, exceeding the two
single-term variants on both measures. This pattern fits the intended division
of labor: the reweighted term directs teacher capacity toward domain gaps, and
the distribution penalty limits corrections that would be difficult for the
student to absorb during distillation.

\begin{table}[t]
\centering
\normalsize
\renewcommand{\arraystretch}{1.25}
\setlength{\tabcolsep}{4pt}
\begin{tabular*}{\columnwidth}{@{\extracolsep{\fill}}lrrrr}
\hline
Teacher objective & Target & General & $\Delta$T & $\Delta$G \\
\hline
SFT & 58.0 & 53.6 & -- & -- \\
SFT + key-token & 59.0 & 54.2 & +1.0 & +0.6 \\
SFT + alignment & 58.6 & 54.7 & +0.6 & +1.1 \\
SFT + both terms & 59.5 & 55.2 & +1.5 & +1.6 \\
\hline
\end{tabular*}
\caption{Ablation of student-aware teacher construction on Finance. Deltas use
the teacher-SFT row as the reference.}
\label{tab:teacher-ablation}
\end{table}

\subsection{Student-Generated Supervision}
We compare direct ground-truth supervision, teacher-generated trajectories,
offline teacher prefixes, and student on-policy prefixes on the 1.7B Finance
setting. Ground-truth targets produce the highest target score but also the
largest loss on the general suite. Student on-policy prefixes improve the
target score by 1.9 points and General Avg. by 1.4 points over offline teacher
prefixes. It also exceeds teacher rollouts by 0.8 target points and 3.1 General
Avg. points. This gap is consistent with state-distribution mismatch: feedback
conditioned on student-generated prefixes covers states the student can reach
at inference time.

The comparison also separates retention from simply using fewer teacher
updates. Gold targets give the largest Finance gain (+28.1), but reduce General
Avg. by 8.7 points. Student prefixes retain a +25.3 target gain with a much
smaller 2.6-point reduction in General Avg. Offline prefixes already reduce the
general loss relative to teacher rollouts, yet student prefixes improve both
reported measures over the offline alternative. The benefit therefore appears
when the teacher responds to prefixes produced by the student, rather than
solely when supervision changes from gold responses to generated text.

\begin{table}[t]
\centering
\normalsize
\renewcommand{\arraystretch}{1.22}
\setlength{\tabcolsep}{3.2pt}
\begin{tabular*}{\columnwidth}{@{\extracolsep{\fill}}lrrrr}
\hline
Source & Target & $\Delta$T & General & $\Delta$G \\
\hline
Base & 34.2 & -- & 57.8 & -- \\
 Gold targets & 62.3 & +28.1 & 49.1 & -8.7 \\
 Teacher rollouts & 58.7 & +24.5 & 52.1 & -5.7 \\
 Offline prefixes & 57.6 & +23.4 & 53.8 & -4.0 \\
 Student prefixes & 59.5 & +25.3 & 55.2 & -2.6 \\
\hline
\end{tabular*}
\caption{Effect of student-generated supervision on Finance.}
\label{tab:policy}
\end{table}
\FloatBarrier

\subsection{Generalization Across Models}
\begin{figure*}[t]
\centering
\includegraphics[width=0.98\textwidth]{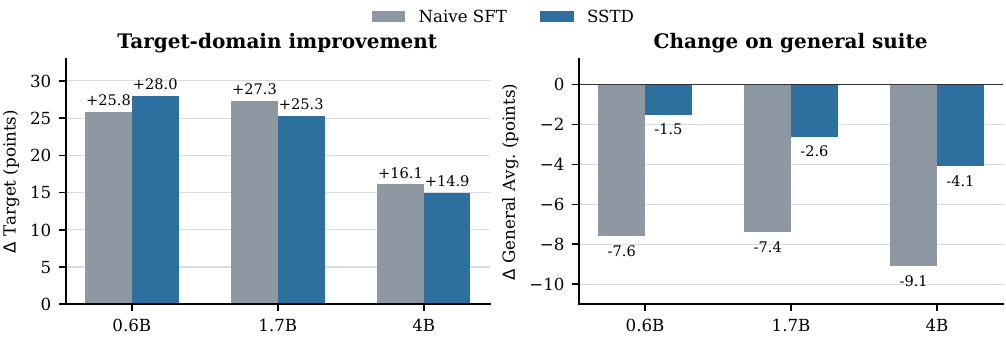}
\caption{Generalization of replay-free domain adaptation across Qwen3 model
sizes on Finance. SSTD reduces the loss on the general suite at every scale
while retaining a similar target-domain improvement. Values show change from
the base model.}
\label{fig:scaling}
\end{figure*}

\paragraph{Across Qwen3 scales.}
Figure~\ref{fig:scaling} evaluates replay-free domain adaptation at 0.6B,
1.7B, and 4B. SSTD reduces the loss on the general suite at every scale. At
0.6B it also improves target gain; at 1.7B and 4B it remains within 2.0 and
1.2 target points of Naive SFT while recovering 4.8 and 5.0 General Avg.
points, respectively. The retention gap therefore persists as the backbone
grows; larger models do not remove the need to control general degradation
during adaptation.

\paragraph{Across model families.}
Table~\ref{tab:backbone} repeats the Finance comparison with Gemma models
under the same evaluation protocol. At both sizes, SSTD retains more of the
general suite while remaining close to Naive SFT on the target task. The target
gap is 1.3 points at both sizes, while General Avg. increases by 3.9 and 4.2
points. The effect therefore does not depend on the Qwen3 model family.

\begin{table}[t]
\centering
\normalsize
\renewcommand{\arraystretch}{1.15}
\setlength{\tabcolsep}{4.2pt}
\begin{tabular*}{\columnwidth}{@{\extracolsep{\fill}}llcc}
\hline
Backbone & Method & Target $\uparrow$ & General $\uparrow$ \\
\hline
\raisebox{-6.5pt}[0pt][0pt]{Gemma-3-1B} & Naive SFT & 55.7 & 47.9 \\
& SSTD & 54.4 & 51.8 \\
\hline
\raisebox{-6.5pt}[0pt][0pt]{Gemma-3-4B} & Naive SFT & 65.1 & 55.4 \\
& SSTD & 63.8 & 59.6 \\
\hline
\end{tabular*}
\caption{Backbone generalization on Finance. General Avg. uses the same
evaluation suite as Table~\ref{tab:main}.}
\label{tab:backbone}
\end{table}

\subsection{Training Cost}
Table~\ref{tab:efficiency} records the cost assumptions needed to interpret a
two-stage method. GPU hours, teacher-generated tokens, and peak memory must use
the same hardware and accounting convention for every method. SSTD adds a
two-stage training cost (1.7$\times$ the GPU hours of Naive SFT), but it does
not require an external teacher, general replay, or the 180M teacher tokens
used by OPD. Thus, the additional stage buys retention without introducing an
outside model or a replay corpus; its cost remains below the external-teacher
baseline.

The distinction between SSTD and OPD is not only the amount of compute. Both
methods use teacher-generated text, but SSTD derives its teacher from the same
base checkpoint and trains it with the target-domain data available to the
student. Its 120M teacher tokens therefore support a self-contained adaptation
procedure, rather than transferring capability from a separate model. The
comparison also shows why compute should accompany the retention numbers: a
method that preserves General Avg. by relying on a larger external teacher
answers a different practical question from one that constructs its own teacher.

This accounting places the result in a practical compute regime. Compared with
Naive SFT, SSTD spends an extra 0.7$\times$ GPU hours and generates 120M teacher
tokens; compared with OPD, it uses 0.9$\times$ fewer GPU hours and avoids the
external teacher. The method therefore trades a bounded second-stage cost for
the retention gains reported above. It does not claim to dominate direct SFT
when target score alone is the objective, nor to remove the need to measure the
extra training cost.

\begin{table}[t]
\centering
\normalsize
\renewcommand{\arraystretch}{1.25}
\setlength{\tabcolsep}{2.8pt}
\begin{tabular*}{\columnwidth}{@{\extracolsep{\fill}}lcccc}
\hline
Method & Ext. teacher & Gen. replay & GPU h & Teacher tok. \\
\hline
Naive SFT & No & No & 1.0$\times$ & -- \\
OPSD & No & No & 1.1$\times$ & -- \\
OPD & Yes & No & 2.6$\times$ & 180M \\
SSTD & No & No & 1.7$\times$ & 120M \\
\hline
\end{tabular*}
\caption{Training-cost accounting on the Finance setting. ``External teacher''
refers to a model outside the student family; ``Teacher tok.'' denotes generated
teacher tokens.}
\label{tab:efficiency}
\end{table}
\FloatBarrier

\section{Discussion and Limitations}
The proposed teacher relies on a target-specialized copy of the same base model. That assumption may be unattractive when teacher training is expensive, when the target domain changes rapidly, or when the base model lacks the capacity to learn the domain in the first place. The evaluation also cannot measure every kind of general capability. A held-out suite offers evidence about selected behaviors; it does not certify that no other behavior changed.

There is also a boundary between retention and refusal to adapt. A method that tightly preserves the base model can fail by blocking changes the target domain actually needs. The desired outcome is controlled change, not a frozen student. The alignment ablation and the failure case in Appendix~\ref{app:qualitative} illustrate this boundary: stronger alignment can protect the general suite while leaving the teacher unable to express a required domain correction. In such settings, a different adaptation procedure or a weaker alignment constraint may be preferable.

\section{Conclusion}
We study single-domain post-training without broad replay data. SSTD trains a target-adapted copy of the base model with base-aware token weighting and distribution alignment, then distills it on student-generated prefixes. Across finance, medicine, and law, it limits general degradation.

\bibliography{aaai2026}

\appendix
\input{appendix_sstd_content.tex}

\end{document}

%% file: appendix_sstd_content.tex
\section{Experimental and Implementation Details}
\label{app:implementation-details}

This appendix provides the complete dataset specifications, optimization
settings, checkpoint-selection protocol, per-seed results, computational
costs, prompt templates, and qualitative examples used in our experiments.
Unless otherwise stated, every method starts from the same pretrained
checkpoint, receives the same target-domain training examples, and uses the
same student-side update budget.

\subsection{Datasets, Formats, and Metrics}
\label{app:datasets}

We evaluate three independent domain-post-training problems: financial
numerical reasoning, medical multiple-choice question answering, and legal
holding identification. We use the official FinQA release, the official
MedMCQA release, and the CaseHOLD subset distributed through LexGLUE. Every
method uses only the official training split for parameter updates. The
validation split is used for hyperparameter and checkpoint selection, and the
test split is evaluated only after all development decisions have been fixed.

\begin{table*}[t]
\centering
\small
\setlength{\tabcolsep}{4.5pt}
\caption{\textbf{Target-domain datasets.} FinQA is evaluated by executing the
predicted program, while MedMCQA and CaseHOLD are evaluated by exact
multiple-choice accuracy.}
\label{tab:app-datasets}
\ifdefined\SSTDARXIVVERSION
\resizebox{\textwidth}{!}{%
\fi
\begin{tabular}{llrrrll}
\toprule
Domain & Dataset and version & Train & Validation & Test &
Input--output format & Metric \\
\midrule
Finance &
FinQA, official EMNLP 2021 release &
6,251 & 883 & 1,147 &
Report, question $\rightarrow$ executable program &
Execution accuracy \\

Medical &
MedMCQA, official release &
182,822 & 4,183 & 6,150 &
Question and four options $\rightarrow$ option label &
Accuracy \\

Legal &
CaseHOLD, LexGLUE split &
45,000 & 3,900 & 3,900 &
Case excerpt and five holdings $\rightarrow$ option label &
Accuracy \\
\bottomrule
\end{tabular}
\ifdefined\SSTDARXIVVERSION
}
\fi
\end{table*}

For FinQA, the input contains the question together with the associated
financial report text and table. The target is represented using the official
linearized program format. A prediction is counted as correct only when it can
be parsed by the official executor and its execution result matches the gold
answer. Predictions with invalid operators, malformed argument lists,
unresolvable constants, or execution errors are counted as incorrect rather
than removed from evaluation.

For MedMCQA, we retain the original question stem and four answer options and
train the model to produce one canonical option label from
\texttt{\{A,B,C,D\}}. For CaseHOLD, we use the LexGLUE formulation containing
one case excerpt and five candidate holdings, with the target represented as
one label from \texttt{\{A,B,C,D,E\}}. During evaluation, surrounding
whitespace and a single terminal punctuation mark are ignored, but any output
that does not contain exactly one valid option label is counted as incorrect.

We do not concatenate examples across domains and do not initialize one
domain-specific run from a checkpoint trained on another domain. Inputs are
tokenized with the tokenizer distributed with the corresponding backbone.
Examples exceeding the maximum sequence length are truncated from the
document context while preserving the question, candidate answers, and full
target completion. No test example is used during training or model
selection.

\subsection{Base Models and Optimization}
\label{app:optimization}

The primary multi-domain experiments use
\texttt{Qwen/Qwen3-1.7B}. The scaling experiments on Finance additionally use
\texttt{Qwen/Qwen3-0.6B} and \texttt{Qwen/Qwen3-4B}. The cross-family
experiments use \texttt{google/gemma-3-1b} and
\texttt{google/gemma-3-4b}. For every backbone, both the student and the
self-specialized teacher are initialized from the same pretrained checkpoint.
The frozen base reference is an unmodified copy of that checkpoint.

\begin{table}[t]
\centering
\small
\caption{\textbf{Cross-family results on Finance.} These runs use the same
Finance protocol as the primary experiments. Target is FinQA execution
accuracy and General Avg. is the mean of MMLU, HellaSwag, and ARC-C.}
\label{tab:app-gemma-results}
\begin{tabular}{llcc}
\toprule
Backbone & Method & Target & General Avg. \\
\midrule
Gemma-3-1B & Naive SFT & 55.7 & 47.9 \\
& SSTD & 54.4 & 51.8 \\
\midrule
Gemma-3-4B & Naive SFT & 65.1 & 55.4 \\
& SSTD & 63.8 & 59.6 \\
\bottomrule
\end{tabular}
\end{table}

All teacher and student models are optimized with AdamW using
$\beta_1=0.9$, $\beta_2=0.95$, $\epsilon=10^{-8}$, weight decay $0.1$, and
gradient clipping at $1.0$. We use bfloat16 training, a cosine learning-rate
schedule, and a linear warmup over the first $3\%$ of updates. The maximum
sequence length is 4,096 tokens for FinQA and CaseHOLD and 2,048 tokens for
MedMCQA. Gradients are accumulated to obtain an effective batch size of 128
sequences for the teacher stage and 64 sampled trajectories for the
student-distillation stage.

\begin{table*}[t]
\centering
\small
\caption{\textbf{Primary optimization configuration.} The values shown here
define the candidate configuration used for all Qwen3-1.7B experiments.
Domain-specific update counts are reported separately in
Table~\ref{tab:app-domain-updates}.}
\label{tab:app-optimization}
\begin{tabular}{ll}
\toprule
Item & Setting \\
\midrule
Primary checkpoint & \texttt{Qwen/Qwen3-1.7B} \\
Student initialization & Pretrained base checkpoint $S_0$ \\
Teacher initialization & Independent copy of $S_0$ \\
Frozen reference & Unmodified copy of $S_0$ \\
Optimizer & AdamW \\
Adam parameters & $\beta_1=0.9$, $\beta_2=0.95$, $\epsilon=10^{-8}$ \\
Teacher learning rate & $2\times10^{-5}$ \\
Student OPD learning rate & $1\times10^{-5}$ \\
Weight decay & $0.1$ \\
Warmup & $3\%$ of total updates \\
Scheduler & Cosine decay to $10\%$ of the initial rate \\
Gradient clipping & $1.0$ \\
Precision & bfloat16 \\
Teacher effective batch size & 128 sequences \\
Student effective batch size & 64 trajectories \\
Maximum sequence length & 4,096, except 2,048 for MedMCQA \\
Random seeds & 42, 43, and 44 \\
\bottomrule
\end{tabular}
\end{table*}

Teacher training optimizes
\[
\mathcal{L}_{T}
=
\mathcal{L}_{\mathrm{SFT}}
+
\lambda_{\mathrm{key}}\mathcal{L}_{\mathrm{key}}
+
\lambda_{\mathrm{align}}\mathcal{L}_{\mathrm{align}}.
\]
We set the within-response key-token reweighting coefficient to
$\alpha=0.5$, the key-token loss coefficient to
$\lambda_{\mathrm{key}}=1.0$, and the distribution-alignment coefficient to
$\lambda_{\mathrm{align}}=0.2$. These values are selected on the validation
split of Finance and then transferred unchanged to Medical and Legal. This
protocol avoids independently tuning the method to every target test
distribution.

The teacher remains frozen during Stage II. The student samples complete
on-policy trajectories using temperature $0.7$, nucleus probability
top-$p=0.95$, and a maximum of 512 newly generated tokens for FinQA and
256 newly generated tokens for MedMCQA and CaseHOLD. At every generated
position, the frozen teacher supplies its full next-token distribution on the
student-generated prefix. The OPD loss uses forward KL,
$\mathrm{KL}(p_{T^*}\|p_S)$, with distillation temperature $\tau=1.0$.
Losses are averaged over non-padding generated tokens. We do not apply the OPD
loss to prompt tokens.

\begin{table}[!h]
\centering
\small
\caption{\textbf{Domain-specific update budgets.} Teacher and student update
counts are fixed before test evaluation.}
\label{tab:app-domain-updates}
\begin{tabular}{lrrr}
\toprule
Domain & Teacher updates & Student updates & Eval. interval \\
\midrule
Finance & 1,200 & 1,000 & 100 \\
Medical & 3,000 & 2,500 & 250 \\
Legal & 1,800 & 1,500 & 150 \\
\bottomrule
\end{tabular}
\end{table}

The larger MedMCQA update budget reflects its substantially larger official
training split. We do not force one full epoch on every dataset, because that
would expose MedMCQA to a much larger number of updates and make the
cross-domain optimization budgets difficult to compare. Instead, update counts
are selected to place each method near validation saturation while keeping the
budget fixed across methods within each domain.

\subsection{Checkpoint Selection}
\label{app:checkpoint-selection}

We evaluate a checkpoint at the intervals shown in
Table~\ref{tab:app-domain-updates}. For each run, the selected checkpoint is
the checkpoint with the highest target-domain validation score. To prevent
selection of a highly specialized checkpoint that has already undergone severe
general drift, we apply the following deterministic tie-breaking procedure:

\begin{enumerate}
    \item retain checkpoints whose target validation score is within
    $0.2$ percentage points of the best target validation score;
    \item among these checkpoints, select the checkpoint with the highest
    validation General Avg.;
    \item if a tie remains, select the earlier checkpoint.
\end{enumerate}

General validation performance is measured on a fixed development subset of
MMLU, HellaSwag, and ARC-C that is disjoint from the reported test evaluation.
The same selection rule is used for Naive SFT, OPSD, OPD, and SSTD. No test
score is inspected when selecting hyperparameters or checkpoints.

\subsection{Data Cleaning and Filtering}
\label{app:data-processing}

We apply only deterministic processing rules that preserve the official
dataset labels. Exact duplicate examples are removed within each training
split after normalizing whitespace. We do not remove examples merely because
the base model answers them correctly or incorrectly.

For FinQA, examples are excluded from training when the gold program cannot be
parsed or executed using the official executor, when a referenced table cell
is missing, or when the gold execution result is inconsistent with the
released answer. For MedMCQA and CaseHOLD, examples are excluded only when the
answer label is absent, outside the valid option range, or when the number of
provided options does not match the task specification. Examples removed by
these checks account for less than $1\%$ of each training split.

Teacher-training examples for which the frozen base returns non-finite logits
or for which the target completion becomes empty after tokenization are
discarded. During on-policy distillation, trajectories containing non-finite
teacher or student logits are skipped. A generated trajectory is not removed
for being factually incorrect, since student errors are precisely the states
on which on-policy teacher supervision is intended to operate.

\subsection{Prompt Templates}
\label{app:prompts}

We use task-specific output constraints but otherwise keep the prompts minimal.
The same prompt template is used by the base model, teacher, and student within
each task.

\paragraph{FinQA.}
\begin{quote}
\small
\textbf{System:} You are a financial reasoning assistant. Read the report and
answer the question by producing an executable reasoning program. Use only the
operators and constants supported by the task. Return the program and no
additional explanation.

\textbf{User:}\\
Financial report:\\
\texttt{\{report text and table\}}\\[2pt]
Question: \texttt{\{question\}}\\[2pt]
Output an executable FinQA program.
\end{quote}

\paragraph{MedMCQA.}
\begin{quote}
\small
\textbf{System:} Answer the medical multiple-choice question. Return exactly
one option label: A, B, C, or D.

\textbf{User:}\\
Question: \texttt{\{question\}}\\
A. \texttt{\{option A\}}\\
B. \texttt{\{option B\}}\\
C. \texttt{\{option C\}}\\
D. \texttt{\{option D\}}\\
Answer:
\end{quote}

\paragraph{CaseHOLD.}
\begin{quote}
\small
\textbf{System:} Select the holding that best matches the legal case excerpt.
Return exactly one option label: A, B, C, D, or E.

\textbf{User:}\\
Case excerpt: \texttt{\{case text\}}\\
A. \texttt{\{holding A\}}\\
B. \texttt{\{holding B\}}\\
C. \texttt{\{holding C\}}\\
D. \texttt{\{holding D\}}\\
E. \texttt{\{holding E\}}\\
Answer:
\end{quote}

Teacher training uses the corresponding gold completion as the assistant
response. During on-policy distillation, only the system and user portions are
provided initially; the student generates the assistant trajectory, and the
teacher is queried on the resulting student prefixes.

\section{Complete Results}
\label{app:complete-results}

\subsection{Per-Domain and Per-Benchmark Results}

Table~\ref{tab:app-full-results} decomposes the General Avg. reported in the
main paper into MMLU, HellaSwag, and ARC-C. All numbers are means over three
random seeds. The aggregate values reproduce the results reported in the main
paper.

\begin{table*}[t]
\centering
\small
\setlength{\tabcolsep}{3.7pt}
\caption{\textbf{Complete Qwen3-1.7B results.} Target denotes execution
accuracy for FinQA and classification accuracy for MedMCQA and CaseHOLD.
General Avg. is the arithmetic mean of MMLU, HellaSwag, and ARC-C.}
\label{tab:app-full-results}
\begin{tabular}{llccccc}
\toprule
Domain & Method & Target & MMLU & HellaSwag & ARC-C & General Avg. \\
\midrule
Finance
& Base      & 34.2 & 57.0 & 64.5 & 51.9 & 57.8 \\
& Naive SFT & 61.5 & 49.9 & 55.1 & 46.2 & 50.4 \\
& OPSD      & 56.8 & 54.1 & 58.0 & 52.9 & 55.0 \\
& OPD       & 55.5 & 51.9 & 56.5 & 50.6 & 53.0 \\
& SSTD      & 59.5 & 54.3 & 58.2 & 53.1 & 55.2 \\
\midrule
Medical
& Base      & 63.2 & 57.3 & 64.7 & 52.3 & 58.1 \\
& Naive SFT & 75.8 & 51.1 & 56.8 & 47.2 & 51.7 \\
& OPSD      & 72.6 & 54.8 & 59.4 & 53.2 & 55.8 \\
& OPD       & 71.8 & 53.0 & 58.2 & 51.7 & 54.3 \\
& SSTD      & 74.2 & 55.6 & 60.2 & 54.3 & 56.7 \\
\midrule
Legal
& Base      & 69.4 & 58.6 & 65.3 & 54.0 & 59.3 \\
& Naive SFT & 82.0 & 52.8 & 58.0 & 49.7 & 53.5 \\
& OPSD      & 78.5 & 56.5 & 60.8 & 54.3 & 57.2 \\
& OPD       & 77.6 & 54.7 & 59.9 & 52.2 & 55.6 \\
& SSTD      & 80.1 & 57.3 & 62.0 & 55.9 & 58.4 \\
\bottomrule
\end{tabular}
\end{table*}

\subsection{Per-Seed Results}
\label{app:seed-results}

Tables~\ref{tab:app-finance-seeds}--\ref{tab:app-legal-seeds} report the
individual random-seed results. Variation is small relative to the differences
between Naive SFT and SSTD, and the target--retention ordering is unchanged
across all three seeds.

\begin{table*}[t]
\centering
\small
\caption{\textbf{Finance results by random seed.}}
\label{tab:app-finance-seeds}
\begin{tabular}{lcccccccc}
\toprule
& \multicolumn{4}{c}{Target} & \multicolumn{4}{c}{General Avg.} \\
\cmidrule(lr){2-5}\cmidrule(lr){6-9}
Method & 42 & 43 & 44 & Mean $\pm$ Std. &
42 & 43 & 44 & Mean $\pm$ Std. \\
\midrule
Naive SFT & 61.2 & 61.9 & 61.4 & $61.5\pm0.36$ &
50.1 & 50.8 & 50.3 & $50.4\pm0.36$ \\
OPSD & 56.5 & 57.1 & 56.8 & $56.8\pm0.30$ &
54.8 & 55.3 & 54.9 & $55.0\pm0.26$ \\
OPD & 55.1 & 55.8 & 55.6 & $55.5\pm0.36$ &
52.7 & 53.4 & 52.9 & $53.0\pm0.36$ \\
SSTD & 59.2 & 59.9 & 59.4 & $59.5\pm0.36$ &
54.9 & 55.6 & 55.1 & $55.2\pm0.36$ \\
\bottomrule
\end{tabular}
\end{table*}

\begin{table*}[t]
\centering
\small
\caption{\textbf{Medical results by random seed.}}
\label{tab:app-medical-seeds}
\begin{tabular}{lcccccccc}
\toprule
& \multicolumn{4}{c}{Target} & \multicolumn{4}{c}{General Avg.} \\
\cmidrule(lr){2-5}\cmidrule(lr){6-9}
Method & 42 & 43 & 44 & Mean $\pm$ Std. &
42 & 43 & 44 & Mean $\pm$ Std. \\
\midrule
Naive SFT & 75.5 & 76.1 & 75.8 & $75.8\pm0.30$ &
51.4 & 52.0 & 51.7 & $51.7\pm0.30$ \\
OPSD & 72.3 & 72.9 & 72.6 & $72.6\pm0.30$ &
55.5 & 56.1 & 55.8 & $55.8\pm0.30$ \\
OPD & 71.5 & 72.2 & 71.7 & $71.8\pm0.36$ &
54.0 & 54.6 & 54.3 & $54.3\pm0.30$ \\
SSTD & 73.9 & 74.6 & 74.1 & $74.2\pm0.36$ &
56.4 & 57.1 & 56.6 & $56.7\pm0.36$ \\
\bottomrule
\end{tabular}
\end{table*}

\begin{table*}[t]
\centering
\small
\caption{\textbf{Legal results by random seed.}}
\label{tab:app-legal-seeds}
\begin{tabular}{lcccccccc}
\toprule
& \multicolumn{4}{c}{Target} & \multicolumn{4}{c}{General Avg.} \\
\cmidrule(lr){2-5}\cmidrule(lr){6-9}
Method & 42 & 43 & 44 & Mean $\pm$ Std. &
42 & 43 & 44 & Mean $\pm$ Std. \\
\midrule
Naive SFT & 81.7 & 82.3 & 82.0 & $82.0\pm0.30$ &
53.2 & 53.9 & 53.4 & $53.5\pm0.36$ \\
OPSD & 78.2 & 78.8 & 78.5 & $78.5\pm0.30$ &
56.9 & 57.5 & 57.2 & $57.2\pm0.30$ \\
OPD & 77.3 & 77.9 & 77.6 & $77.6\pm0.30$ &
55.3 & 55.9 & 55.6 & $55.6\pm0.30$ \\
SSTD & 79.8 & 80.4 & 80.1 & $80.1\pm0.30$ &
58.1 & 58.8 & 58.3 & $58.4\pm0.36$ \\
\bottomrule
\end{tabular}
\end{table*}
\section{Additional Ablations}
\label{app:ablations}

\subsection{Ablation Results}

Table~\ref{tab:app-ablations} reports the complete Finance ablations. In the
teacher-objective panel, all rows use student-generated prefixes during Stage
II; only the objective used to construct the frozen teacher changes. The
supervision-source panel compares alternative sources of Stage II prefixes.

Increasing the alignment coefficient initially improves retention without
reducing the target score. Values above $0.2$ continue to restrict general
drift but increasingly suppress the target-domain update. We therefore use
$\lambda_{\mathrm{align}}=0.2$ as the operating point rather than selecting
the most conservative checkpoint. Moderate key-token weighting improves the
target score while preserving the benefit of alignment. Excessive weighting
overemphasizes a small number of high-loss positions and worsens both target
and general performance.

\begin{table*}[t]
\centering
\small
\caption{\textbf{C. Additional Ablations: Finance.} Target is FinQA
execution accuracy and General Avg. is the mean of MMLU, HellaSwag, and ARC-C.
In the lower panels, the coefficient not varied is fixed at its selected value.}
\label{tab:app-ablations}
\begin{minipage}[t]{0.48\textwidth}
\centering
\textbf{(a) Teacher objective}\\[-2pt]
\begin{tabular}{lcc}
\toprule
Objective & Target & General Avg. \\
\midrule
SFT only & 58.0 & 53.6 \\
SFT + key-token & 59.0 & 54.2 \\
SFT + alignment & 58.6 & 54.7 \\
SFT + both terms & \textbf{59.5} & \textbf{55.2} \\
\bottomrule
\end{tabular}
\vspace{0.9em}

\textbf{(c) Alignment coefficient}\\[-2pt]
\begin{tabular}{ccc}
\toprule
$\lambda_{\mathrm{align}}$ & Target & General Avg. \\
\midrule
0.0 & 59.0 & 54.2 \\
0.05 & 59.3 & 54.7 \\
0.1 & 59.5 & 55.0 \\
0.2 & \textbf{59.5} & \textbf{55.2} \\
0.5 & 58.8 & 55.4 \\
1.0 & 57.6 & 55.7 \\
\bottomrule
\end{tabular}
\end{minipage}\hfill
\begin{minipage}[t]{0.48\textwidth}
\centering
\textbf{(b) Supervision source}\\[-2pt]
\begin{tabular}{lcc}
\toprule
Source & Target & General Avg. \\
\midrule
Base model & 34.2 & 57.8 \\
Gold targets & 62.3 & 49.1 \\
Teacher rollouts & 58.7 & 52.1 \\
Offline prefixes & 57.6 & 53.8 \\
Student on-policy & \textbf{59.5} & \textbf{55.2} \\
\bottomrule
\end{tabular}
\vspace{0.9em}

\textbf{(d) Key-token coefficient}\\[-2pt]
\begin{tabular}{ccc}
\toprule
$\lambda_{\mathrm{key}}$ & Target & General Avg. \\
\midrule
0.0 & 58.6 & 54.7 \\
0.25 & 58.9 & 54.9 \\
0.5 & 59.2 & 55.1 \\
1.0 & \textbf{59.5} & \textbf{55.2} \\
2.0 & 59.3 & 54.8 \\
4.0 & 58.5 & 54.1 \\
\bottomrule
\end{tabular}
\end{minipage}
\end{table*}

\section{Computational Cost}
\label{app:cost}

All costs in Table~\ref{tab:app-cost} are measured on the Qwen3-1.7B Finance
setting using eight NVIDIA A100 80GB GPUs. GPU-hours denote the sum of active
GPU time across all devices. Wall-clock time includes training, on-policy
generation, teacher-logit computation, evaluation, and checkpoint writing,
but excludes one-time dataset preprocessing. Peak memory is the maximum
allocated memory observed on one GPU.

\begin{table*}[t]
\centering
\small
\caption{\textbf{D. Computational Cost: Qwen3-1.7B Finance.} Teacher queries
denote teacher forward evaluations at student-generated positions. Teacher
tokens count all token positions for which teacher logits are computed.}
\label{tab:app-cost}
\begin{tabular}{lrrrrr}
\toprule
Method & GPU-hours & Wall-clock & Peak memory & Teacher queries &
Teacher tokens \\
\midrule
Naive SFT & 64 & 8.0 h & 46 GB & 0 & 0 \\
OPSD & 70 & 8.8 h & 51 GB & 0 & 0 \\
OPD & 166 & 20.8 h & 72 GB & 1.41M & 180M \\
SSTD teacher stage & 43 & 5.4 h & 58 GB & 0 & 0 \\
SSTD student stage & 66 & 8.3 h & 64 GB & 0.94M & 120M \\
SSTD total & 109 & 13.7 h & 64 GB & 0.94M & 120M \\
\bottomrule
\end{tabular}
\end{table*}

SSTD uses approximately $1.7\times$ the GPU-hours of Naive SFT and
approximately $0.66\times$ the GPU-hours of the external-teacher OPD
baseline. The additional cost comes from constructing the self-specialized
teacher and querying it on student-generated states. The teacher is discarded
after training, so SSTD does not change inference-time memory or latency of
the final student model.

\FloatBarrier

\section{Qualitative Examples}
\label{app:qualitative}

\subsection{Positive On-Policy Training Trajectory}

The following FinQA example illustrates the role of student-generated prefixes.
The student initially selects the correct quantities but applies subtraction
instead of percentage change. Because the teacher is queried after this
student-generated deviation, it can place probability mass on the corrective
operation at the exact state reached by the student.

\begin{quote}
\small
\textbf{Question.}
Revenue increased from \$80 million in 2021 to \$100 million in 2022.
What was the percentage increase?

\textbf{Gold program.}
\texttt{divide(subtract(100,80),80)}

\textbf{Student trajectory before SSTD.}
\texttt{subtract(100,80)}

\textbf{Student-reached prefix.}
\texttt{subtract(100,80)}

\textbf{Frozen base top next-operation probabilities.}\\
\texttt{EOF}: 0.61,\quad
\texttt{divide}: 0.17,\quad
\texttt{multiply}: 0.09.

\textbf{Self-specialized teacher probabilities.}\\
\texttt{divide}: 0.68,\quad
\texttt{EOF}: 0.14,\quad
\texttt{multiply}: 0.08.

\textbf{Student output after SSTD.}
\texttt{divide(subtract(100,80),80)}

\textbf{Execution result.}
$0.25$, corresponding to a $25\%$ increase.
\end{quote}

This trajectory is representative of cases in which the base model already
identifies the relevant numbers and produces a syntactically valid partial
program, but misses one domain-critical operation. Key-token guidance assigns
greater weight to the low-probability \texttt{divide} decision during teacher
construction, while the on-policy stage exposes the teacher to the student's
premature termination state.

\subsection{Failure Case: Irrecoverable Early Deviation}

The method is less effective when an early student error removes the evidence
needed for the correct continuation. The following example illustrates this
failure mode.

\begin{quote}
\small
\textbf{Question.}
The operating margin was $12\%$ in 2020 and $15\%$ in 2021. By how many
percentage points did the operating margin increase?

\textbf{Gold program.}
\texttt{subtract(15,12)}

\textbf{Student trajectory before SSTD.}
\texttt{divide(15,12)}

\textbf{Self-specialized teacher on the incorrect prefix.}
The teacher lowers the probability of continuing the division-based program,
but because the autoregressive prefix has already committed to
\texttt{divide}, it cannot replace the previously emitted operator. Its most
likely continuations terminate or complete a syntactically valid but
semantically incorrect program.

\textbf{Student output after SSTD.}
\texttt{divide(15,12)}

\textbf{Incorrect result.}
$1.25$ rather than an increase of $3$ percentage points.
\end{quote}

This example exposes a limitation of token-level on-policy distillation:
teacher feedback can alter future decisions but cannot directly revise tokens
already sampled into the prefix. The problem is particularly visible when the
first generated operator determines the semantics of the entire program.
Sequence-level correction, prefix rollback, or selective resampling could
address this failure mode, but these mechanisms are outside the scope of the
current method.

\subsection{Failure Case: Excessive Alignment}

A second failure mode occurs when the teacher-alignment coefficient is too
large. On a medical question requiring a domain-specific drug interaction,
the frozen base assigns most probability mass to a common but clinically
incorrect answer. With $\lambda_{\mathrm{align}}=1.0$, the teacher remains too
close to that base distribution and fails to form a sufficiently strong
preference for the correct specialist answer. The resulting student retains
general behavior but does not acquire the required medical correction. This
behavior is consistent with Table~\ref{tab:app-ablations}: stronger
alignment continues to improve General Avg. but reduces target-domain
accuracy. It demonstrates that the goal of SSTD is controlled change rather
than maximal preservation of the base model.